\documentclass{article}

\usepackage{PRIMEarxiv}

\usepackage[utf8]{inputenc} % allow utf-8 input
\usepackage[T1]{fontenc}    % use 8-bit T1 fonts
\usepackage{hyperref}       % hyperlinks
\usepackage{url}            % simple URL typesetting
\usepackage{booktabs}       % professional-quality tables
\usepackage{amsfonts}       % blackboard math symbols
\usepackage{nicefrac}       % compact symbols for 1/2, etc.
\usepackage{microtype}      % microtypography
\usepackage{lipsum}
\usepackage{fancyhdr}       % header
\usepackage{graphicx}       % graphics    

\usepackage{tabularx}
\usepackage{geometry}
\usepackage{tocloft}
\usepackage{titletoc}
\usepackage{mdframed}
\usepackage{makecell}
\usepackage{subcaption}
\usepackage{longtable}
\usepackage{amsmath}
\usepackage{amssymb}
\usepackage{pgf}
\usepackage{multirow}
\usepackage{algorithm}
\usepackage{algorithmic}
\usepackage{bbm}

\DeclareMathOperator{\PE}{PE}

\fancypagestyle{firstpage}{
  \fancyhf{} % limpia todo
  \fancyfoot[L]{{\parbox{\textwidth}{V. Toscano et al., Safe and Efficient Social Navigation through Explainable Safety Regions Based on Topological Features. Proceedings of the 3rd World Conference on eXplainable Artificial Intelligence (XAI 2025), Istambul, Turkey, July 9-11, 2025.
  }}} % <-- aquí cambias tu nota
}

\title{Safe and Efficient Social Navigation through Explainable Safety Regions Based on Topological Features
}
\author{ \href{https://orcid.org/0009-0006-1316-9026}{\includegraphics[scale=0.06]{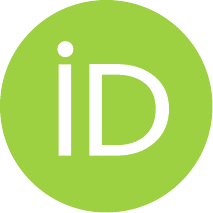}}Victor Toscano-Duran\thanks{Corresponding author.}, \href{https://orcid.org/0000-0001-9937-0033}{\includegraphics[scale=0.06]{orcid.pdf}}Rocio Gonzalez-Diaz \\
  Department of Applied Mathematics I, University of Seville \\
  Seville, Spain \\
  \texttt{\{vtoscano, rogodi\}us.es} \\
   \And
  \href{https://orcid.org/0000-0002-0579-647X}{\includegraphics[scale=0.06]{orcid.pdf}}Sara Narteni, \href{https://orcid.org/0000-0002-7206-5511}{\includegraphics[scale=0.06]{orcid.pdf}}Alberto Carlevaro, \href{https://orcid.org/0000-0001-6201-6225}{\includegraphics[scale=0.06]{orcid.pdf}}Maurizio Mongelli \\
  CNR-IEIIT \\
  Genoa, Italy \\
  \texttt{\{sara.narteni,maurizio.mongelli,albertocarlevaro\}@cnr.it} \\
  \And
  \href{https://orcid.org/0000-0002-1263-4110}{\includegraphics[scale=0.06]{orcid.pdf}}Jerome Guzzi \\
  SUPSI, IDSIA \\
  Lugano, Switzerland \\
  \texttt{jerome.guzzi@supsi.ch}
}

\begin{document}
\maketitle
\thispagestyle{firstpage}

\begin{abstract} 
The recent adoption of artificial intelligence (AI) in robotics has driven the development of algorithms that enable autonomous systems to adapt to complex social environments. In particular, safe and efficient social navigation is a key challenge, requiring AI not only to avoid collisions and deadlocks but also to interact intuitively and predictably with its surroundings. To date, methods based on probabilistic models and the generation of conformal safety regions have shown promising results in defining safety regions with a controlled margin of error, primarily relying on classification approaches and explicit rules to describe collision-free navigation conditions.

This work extends the existing perspective by investigating how topological features can contribute to the creation of explainable safety regions in social navigation scenarios, enabling the classification and characterization of different simulation behaviors. Rather than relying on behaviors parameters to generate safety regions, we leverage topological features through topological data analysis. We first utilize global rule-based classification to provide interpretable characterizations of different simulation behaviors, distinguishing between safe (free of collisions) and unsafe scenarios based on topological properties. Next, we define safety regions, $S_\varepsilon$, representing zones in the topological feature space where collisions are avoided with a maximum classification error of $\varepsilon$. These regions are constructed using adjustable SVM classifiers and order statistics, ensuring a robust and scalable decision boundary. To enhance interpretability, we extract local rules from these safety regions, ensuring that the decision-making process remains transparent and comprehensible.

Initially, we generate safety regions that separate simulations with and without collisions, achieving higher accuracy than methods that do not incorporate topological features. This approach also provides a deeper and more intuitive understanding of robot interactions within a navigable space. We then extend our methodology to design safety regions that ensure efficient simulations (i.e., free of deadlocks). Finally, we integrate both aspects to obtain comprehensive safety regions that guarantee both collision-free and deadlock-free simulations, defining an overall compliant simulation space.

\keywords{Safe navigation \and Explainable 
artificial intelligence \and Safety regions \and Topological data analysis \and Interpretability}

\end{abstract}

\section{Introduction}

Nowadays, machine learning (ML) and deep learning (DL) play a fundamental role in a wide range of fields \cite{rao2024artificial}, with robotics standing out as a domain where autonomous decision-making finds fertile ground. Artificial intelligence (AI) supports a wide range of robotics applications \cite{vrontis2023artificial}, from object detection and recognition to healthcare, manufacturing, and agriculture. One of the most impactful applications of AI in robotics is assistive robotics, where social robots are designed to support everyday tasks, ensuring safe indoor and outdoor navigation while interacting effectively with humans and their surroundings.

However, as autonomous systems become more prevalent in social spaces, ensuring their safety and reliability remains a critical challenge \cite{koopman2017autonomous}. AI models must operate as intended without harming users or the surrounding environment. To address this, many researchers in safe and explainable AI (XAI) \cite{longo2024explainable} have focused on simulation and validation techniques to assess model reliability.

In particular, topological data analysis (TDA) has emerged as a promising tool in this context, offering new ways to characterize interactions and structures within the navigation space \cite{garcia2015vision,islam2020framework}. Within TDA, persistent entropy \cite{rucco2017new} stands out for its ability to efficiently track the evolution of topological features over time, and can be a robust measure for defining safe and efficient navigation regions and analyzing robot-agent interactions.

In this work, we address a simulated social navigation scenario inspired by human movement, where robots navigate between pairs of opposing targets, facing potential risks of collision and deadlock. These challenges compromise both the safety and efficiency of mobile navigation. To tackle this issue, we leverage topological features, specifically from persistent entropy, as input data to classify and characterize different simulation behaviors.

In a fleet of robots, topological features provide a quantifiable representation of the spatial structures that emerge during navigation, capturing patterns such as clustering, dispersion, and movement coordination. This allows us to systematically distinguish between safe (collision-free) and unsafe scenarios. By utilizing global rule-based classification, adjustable SVM classifiers, and order statistics, we construct safety regions that not only prevent collisions but also enhance navigation efficiency by avoiding deadlocks. Finally, we integrate these aspects to define a compliant navigation space, ensuring both collision-free and deadlock-free simulations. This approach supports the development of robust and explainable AI models, where topological insights can enhance transparency and performance in defining safe (without collisions), efficient (without deadlocks), and compliant (free of collisions and deadlocks) navigation strategies.

The remainder of the article is organized as follows: first, in Section \ref{sec:relatedwork}, we present the state of the art and existing literature in the field. Section \ref{sec:background} introduces the preliminary concepts relevant to this study, including topology, confidence regions, and rule-based models, as well as the simulation-based robotic navigation environment. Section \ref{sec:methodology} details the proposed methodology for constructing safety regions regarding the topological feature space. Section \ref{sec:experiments} provides a comprehensive explanation of the experiments conducted, along with the results obtained. Finally, conclusions and future work are discussed in Section \ref{sec:conclusion}.

\section{Related work}\label{sec:relatedwork}

The integration of machine learning (ML) and deep learning (DL) into robotics \cite{murphy2019introduction} has accelerated the capabilities of autonomous systems across various fields. Particularly within assistive and social robotics, the objective has evolved from simple navigation to enabling safe and interpretable decision-making around people. Early works on ML for robotics primarily focused on object detection and recognition \cite{Pierson2017DeepLI,DLrobotics}, yet more recent efforts leverage DL algorithms to enhance contextual understanding and adaptive navigation. These advancements have led to behavioral AI models capable of dynamically adjusting to complex, changing environments, such as crowded indoor spaces while ensuring user safety.

The concept of explainable AI (XAI) \cite{dwivedi2023explainable,saeed2023explainable,longo2024explainable} has emerged as a necessary framework to improve AI transparency and trustworthiness \cite{kaur2022trustworthy}, particularly in safety-critical applications like social robotics. In this context, several studies propose the use of simulation and validation techniques to evaluate ML models before deployment in real-world scenarios, emphasizing safety and reliability in decision-making \cite{anjomshoae2019explainable,atakishiyev2024explainable}. Additionally, explainability and interpretability enhance user trust, as evidenced in approaches like rule-based reinforcement learning and fuzzy logic systems for robotic risk mitigation, which incorporate predefined standards and probabilistic safety regions to guide exploration while minimizing risk and redundant actions \cite{carlevaro2023probabilistic,10.1007/978-3-031-63803-9_22}.

To address this, topological data analysis (TDA) provides novel tools for examining complex data structures, making it possible to extract and interpret information about the underlying structure of navigational spaces. TDA techniques, such as persistent homology \cite{hatcher2005algebraic,carlsson2020topological,edelsbrunner2022computational}, have been applied to analyze spatial connectivity and the evolving structures within navigation environments \cite{TopMapsRobotNavg,ESTEVE2024101953,TopoNav}, offering a quantifiable approach to characterize spatial and behavioral patterns over time. Persistent entropy \cite{chintakunta2015entropy}, a summarization of persistent homology, quantifies the complexity of topological features over time, and is especially useful in scenarios where robots may need to navigate shared spaces while avoiding collisions and deadlocks to ensure safe and efficient navigations.

Moreover, persistent entropy has demonstrated its effectiveness in a variety of other applications such as characterizing idiotypic immune networks \cite{rucco2016characterisation}, analyzing similarities in piecewise linear functions and time series \cite{rucco2017new}, and separating topological features from noise in Vietoris-Rips complexes \cite{atienza2019persistent}. These successful applications underscore its versatility and potential to provide robust insights across diverse domains. Due to its simplicity (providing a single scalar value at each time step making it computationally efficient while still capturing essential topological features), interpretability, stability \cite{atienza2020stability}, and proven success in other applications, as previously mentioned, make it an ideal choice for our work.

To the best of our knowledge, there are no previous works that utilize TDA-based methods specifically for defining safety regions in robot navigation. We have reviewed the state of the art and found that TDA has indeed been explored in various aspects of robot navigation, particularly for characterizing interactions and spatial structures (e.g., \cite{garcia2015vision,islam2020framework,TopMapsRobotNavg,ESTEVE2024101953,TopoNav}). However, none of these works explicitly address the definition of safety regions using TDA.

The integration of TDA into the analysis of robotic systems marks a step forward in developing explainable models for safe navigation. This approach allows us to detect, quantify, and analyze safety regions, focusing on spatial structures formed by robots in motion \cite{lum2013extracting}. By combining TDA with simulated environments, we contribute to the growing field of XAI for robotics by facilitating a more interpretable approach to parameter tuning within navigation models. This directly contributes to improving safety and reliability in socially aware robotic systems, offering the potential for broader applications in real-world, and human-centric environments  \cite{hassija2024interpreting}.

\section{Background}\label{sec:background}

This section presents all preliminary concepts that are used throughout the paper. It begins by introducing topological foundations (Section \ref{sec:topback}), including an exploration of topological spaces and simplicial complexes as fundamental structures. Persistent homology and persistent entropy are also discussed, as they provide insights into the geometry and connectivity of data. Subsequently, the focus shifts to confidence regions (Section \ref{sec:confregback}), defining safety regions based on probabilistic scaling and conformal prediction methods. These techniques ensure robust predictions with statistical guarantees. It continues explaining rule-based models (Section \ref{sec:rulebasedback}), emphasizing interpretability through global and local rule-extraction techniques, such as Anchors. However, these techniques often struggle to capture the underlying geometric and topological structure of data, limiting their ability to generalize across complex decision boundaries. To address this, the proposed methodology in Section \ref{sec:methodology} leverages topological features as inputs to enhance model interpretability and robustness. This subsection also highlights evaluation metrics like coverage and error for assessing rule-based classifiers' performance. To apply these concepts, as well as the methodology proposed in this paper, which is explained in Section \ref{sec:methodology}, to the field of robotic navigation, a social robotics navigation simulator is employed, called Navground (described in Section \ref{sec:navground}). This simulator allows for testing navigation algorithms within multi-agent systems and evaluating their performance in diverse scenarios.

\subsection{Topology Background}\label{sec:topback}

\subsubsection{Topology and Topological Space.}

A topological space is a powerful mathematical concept for describing the connectivity of a space. Informally, a topological space is a set of points, each of them equipped with the notion of neighbourhood. 

One way to represent a topological space is by decomposing it into simple pieces such that their common intersections are lower-dimensional pieces of the same kind. In this paper, we use simplicial complexes as the data structure to represent topological spaces.

\subsubsection{Simplicial Complex.}

An \emph{abstract simplicial complex} $\mathcal{K}$ is given by:
\begin{itemize}
    \item A set $V$ of $0$-simplices (also called vertices);
    \item For each $k \geq 1$, a set of $k$-simplices $\sigma = \{v_0, v_1, \ldots, v_k\}$, where $v_i \in V$;
    \item Each $k$-simplex has $k+1$ faces obtained by removing one of the vertices;
    \item If $\sigma$ belongs to $\mathcal{K}$, then all the faces of $\sigma$ must belong to $\mathcal{K}$.
\end{itemize}

A \emph{simplicial complex} $K$ is a geometric realization of an abstract simplicial complex $\mathcal{K}$ \cite{jonsson2008simplicial}. It is constructed as a nested family of simplices, where each simplex is a generalization of geometric shapes of various dimensions. Specifically, a $0$-simplex corresponds to a point in a given Euclidean space $\mathbb{R}^n$, a $1$-simplex represents an edge, a $2$-simplex is a filled triangle, a $3$-simplex is a filled tetrahedron, and so on. This hierarchical structure allows the simplicial complex to model higher-dimensional relationships between points in a mathematically rigorous way. 

Homology groups are algebraic structures that describe features of a topological space $\mathcal{C}$. The $k$-th Betti number represents the rank of the $k$-dimensional homology group. Informally, for a fixed $k$, the $k$-th Betti number $\beta_k$ counts the number of $k$-dimensional holes characterizing $\mathcal{C}$: $\beta_0$ is the number of connected components, $\beta_1$ counts the number of holes in $2D$ or tunnels in $3D$ \footnote{$dD$ refers to the $d$ - dimensional space $\mathbb{R}^d$}, $\beta_2$ can be thought of as the number of voids in geometric solids. In this work, we use simplicial homology, which consists of computing homology groups of simplicial complexes.

See \cite{hatcher2005algebraic,munkres2018elements} for an introduction to algebraic topology.

\subsubsection{Persistent Homology.}

Topological data analysis, particularly persistent homology, has emerged as a
valuable approach for studying the geometry and connectivity of datasets that
evolve over time or space \cite{lum2013extracting}. The primary objective of employing topological data analysis tools, like persistent homology and persistent entropy, in the context of fleet behavior modeling is to achieve a more nuanced understanding of system dynamics.

Persistent homology is a method for computing $k$-dimensional holes at different spatial resolutions. 
In this section, we briefly explain how this method works.
For a more formal description, we refer the reader to \cite{edelsbrunner2022computational}. 
 
\begin{figure}[h!]
    \centering
    \includegraphics[width=0.8\textwidth,height=0.15\textwidth]{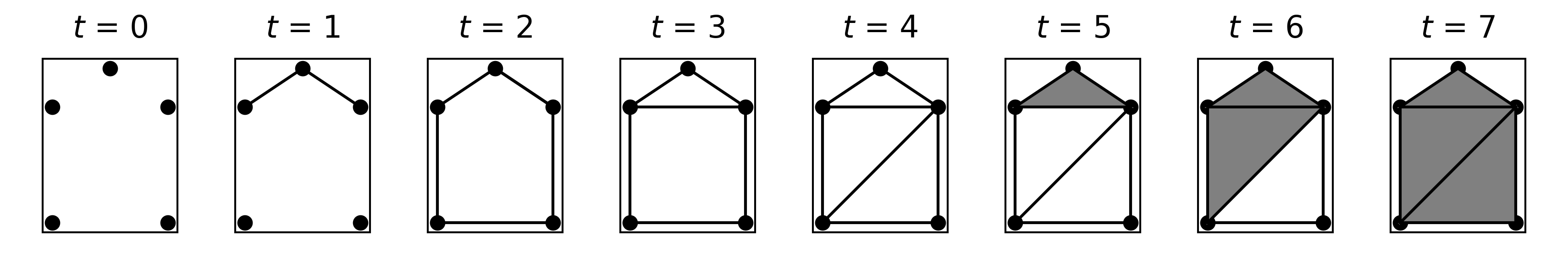}
    \caption{A filtration for $t = 0, 1, 2, 3, 4, 5, 6, 7$ (from left to right).}
    \label{fig:filtration}
\end{figure}

In order to compute persistent homology, we need a nested sequence of increasing subcomplexes. More formally, a \emph{filtered simplicial complex} or, for short, a \emph{filtration} is a collection of subcomplexes $\{K(t) \mid t \in \mathbb{R}\}$ of a simplicial complex $K$ such that $K(t) \subseteq K(s)$ for $t < s$ and there exists $t_{\max} \in \mathbb{R}$ such that $K(t_{\max}) = K$. The \emph{filtration time} (or filter value) of a simplex $\sigma \in K$ is the smallest $t$ such that $\sigma \in K(t)$. An example of a filtered simplicial complex is depicted in Fig.~\ref{fig:filtration}.

Persistent homology describes how the homology of $K$ changes along a filtration $\{K(t) \mid t \in \mathbb{R}\}$. A $k$-dimensional \emph{Betti interval}, with endpoints $[t_{\text{start}}, t_{\text{end}})$, corresponds to a $k$-dimensional hole that appears at filtration time $t_{\text{start}}$ and remains until time $t_{\text{end}}$. 

\begin{figure}[h!]
\centering
\begin{subfigure}{0.4\textwidth}
    \includegraphics[width=1.1\textwidth,height=0.9\textwidth]{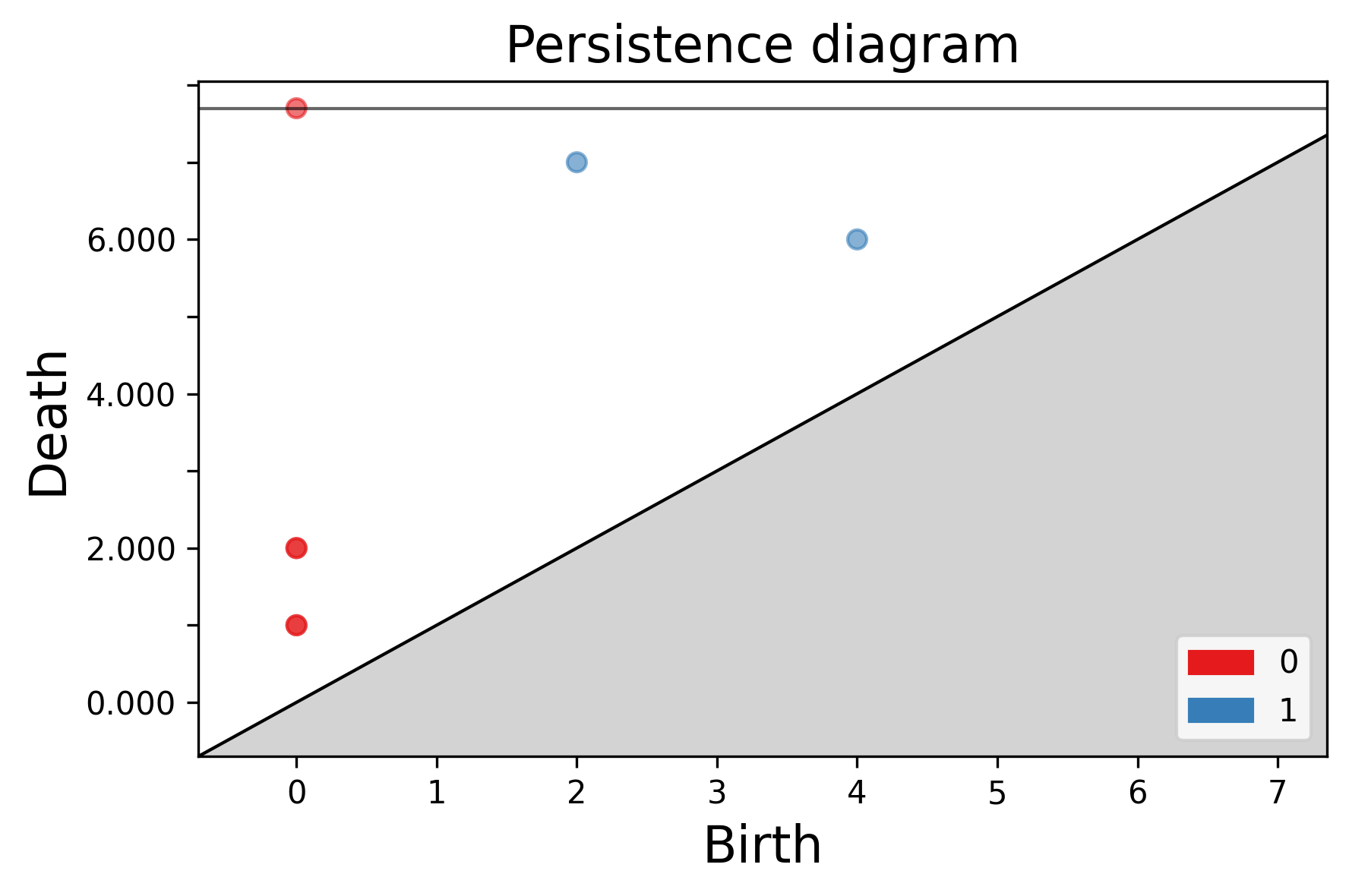}
    \caption{Persistence diagram for filtration in Fig.\ref{fig:filtration}.}
    \label{fig:persistencediagram}
\end{subfigure}
\hfill
\begin{subfigure}{0.4\textwidth}
    \includegraphics[width=1.1\textwidth,height=0.9\textwidth]{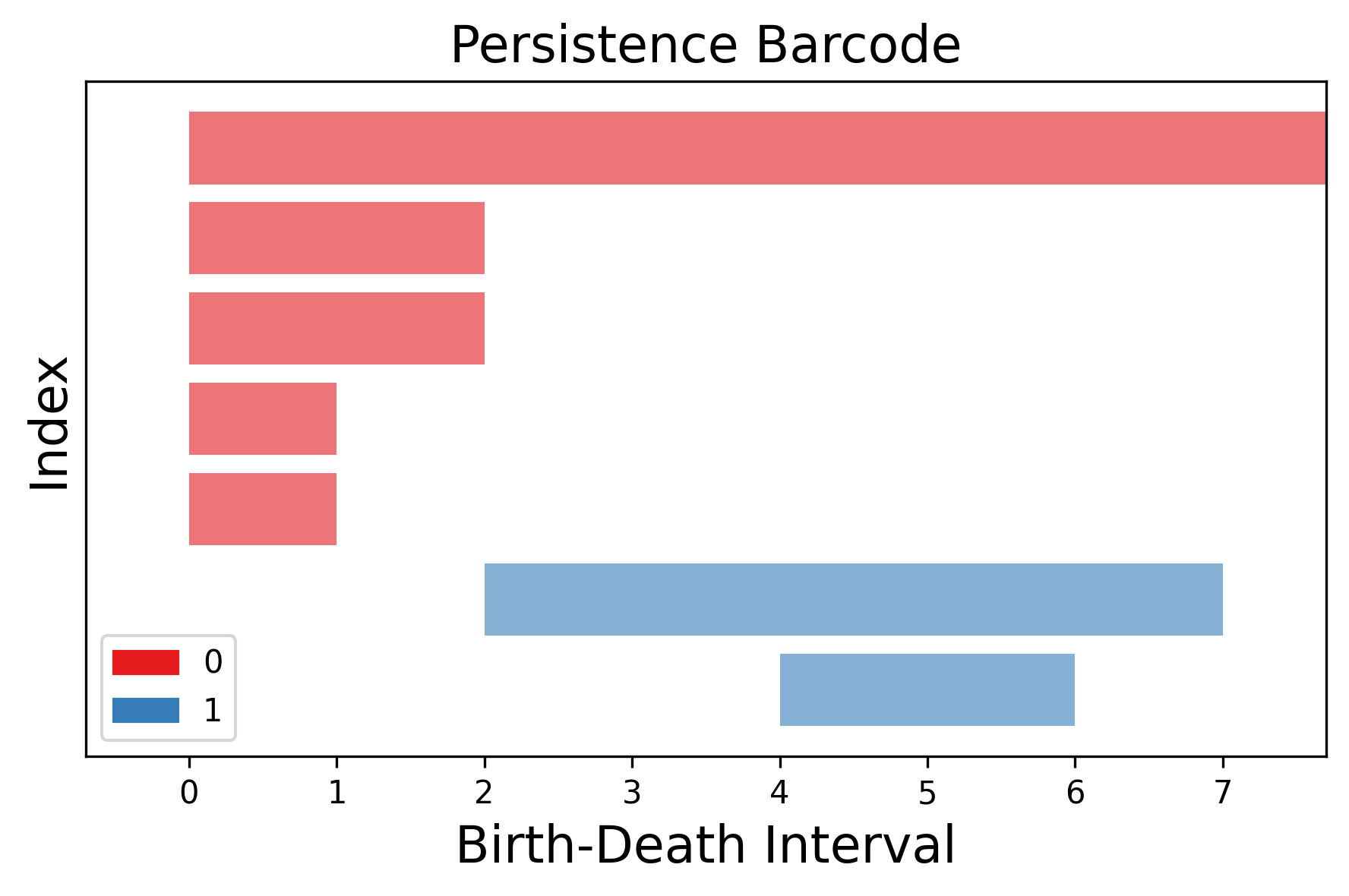}
    \caption{Persistence barcodes for 0-dim and 1-dim for filtration in Fig.\ref{fig:filtration}.}
    \label{fig:persistencebarcode}
\end{subfigure}
\caption{Persistence diagram and persistence barcode example.}
\label{fig:pdandpb}
\end{figure}

To visualize the persistence of topological features, persistence diagrams are commonly used. \textbf{Persistence diagrams} are visual representations used in the study of persistent homology to illustrate the birth and death of topological features (such as connected components, loops, and voids) as the scale of observation changes. It consists in a set of points $(b_i, d_i) \in \mathbb{R}^2$ (with $b_i \leq d_i$). Each point represents the birth ($b_i$) and death ($d_i$) of a topological feature in a filtered space. Instead of points, we sometimes draw intervals representing the birth and death times of homology classes. This method, also known as \textbf{persistence barcodes} \cite{ghrist2008barcodes}, provides an alternative visualization of the same information contained in persistence diagrams, plotting this data in bars. Each bar in the barcode corresponds to a topological feature and is represented by an interval [$b_i, d_i$). An illustrative example of a persistence diagram and a persistence barcode is in Fig. \ref{fig:pdandpb}.

\subsubsection{Persistent Entropy.}

In order to measure how much the construction of a filtered simplicial complex is ordered, an entropy measure, called persistent entropy, was defined in \cite{rucco2017new}. A precursor of this definition was given in \cite{chintakunta2015entropy} to measure how different bars of a persistence barcode are in length. In other words, persistent entropy is a measure of the complexity of a topological space based on its persistence diagram.

Given a filtered simplicial complex $\{K(t) : t \in \mathbb{R}\}$, and the corresponding persistence diagram $D = \{(b_i, d_i) : i \in I\}$, where $b_i \leq d_i$ for all $i \in I$, and $I$ is the index set that identifies the pairs $(b_i,d_i)$ in $D$. As commented previously, each pair $(b_i,d_i)$ corresponds to a topological feature that ``appears'' at time $b_i$ (birth) and ``disappears'' at time $d_i$ (death) as the filtration progresses. The \emph{persistent entropy} $\PE$ of a filtered simplicial complex $K$ is calculated as follows:

\begin{equation}
    \label{eq:entropy}
    \PE = - \sum_{i \in I} p_i \ln{p_i}
\end{equation}
where $p_i = \frac{\ell_i}{L}$, $\ell_i = d_i - b_i$, and $L = \sum_{i \in I} \ell_i$. 

It is important to note that this formulation considers only the duration $\ell_i$ of each topological feature (bar), rather than its starting time $b_i$. This is because persistent entropy aims to capture the distribution of topological feature lifetimes rather than their birth times. Ignoring starting times ensures that the measure remains invariant under time shifts in the filtration, making it more robust for comparing different persistence barcodes. In particular, we focus on dimension 0, which is characterized by the fact that all features in this dimension are born at the same instant but have different durations, meaning some persist longer than others. 

The maximum persistent entropy corresponds to the situation in which all the intervals in the barcode are of equal length. In that case, $\PE = \ln{n}$, where $n$ is the number of elements of $I$ (i.e, the number of bars in the persistence diagram $D$). In contrast, the value of the persistent entropy decreases as more intervals of different lengths are present. A potential point of confusion arises because persistent entropy does not measure the entropy of the distribution of bar lengths directly. Instead, it measures the entropy of the probability distribution induced by the relative contributions of each bar to the total barcode length. If all bars had the same length, their contributions would be equal, maximizing entropy. Conversely, when some bars dominate the total length, entropy decreases. Thus, persistent entropy should be interpreted as the entropy of the distribution of bar indices weighted by their respective lifetimes, rather than the entropy of the length distribution itself.

For example, for the persistence barcode of dimension 0 in Fig. \ref{fig:persistencebarcode}, we have 5 bars (red bars correspond to 0 dimension topological features) with the following lengths $l_i$: $1,1,2,2,8$, obtaining a total length $L$ of $14$. Thus, the probabilities ($p_i$) in this case are: $\frac{1}{14}$, $\frac{1}{14}$, $\frac{2}{14}$, $\frac{2}{14}$, and $\frac{8}{14}$, resulting in a persistent entropy of 1.25.

The stability insights derived from persistence diagrams~\cite{cohen-steiner2007stability} and persistent entropy~\cite{atienza2020stability,rucco2017new} offer a solid foundation for measuring the stability of topological features in response to minor changes. In particular, they guarantee that small perturbations in the input lead to short bars in the persistence barcode and result in minimal, bounded shifts in the persistent entropy. This property is especially useful in contexts like robot navigation, where slight variations in the robot's position—such as those occurring between two time instances with minimal displacement—can otherwise lead to significant uncertainties in the system’s analysis.

\subsection{Background on Confidence Regions}\label{sec:confregback}

Here, the concept of \emph{probabilistic safety region} is introduced, which we define as the subset of the input space where probabilistic guarantees for the prediction of a target (safe) class are provided. This approach begins with a discussion of adjustable classifiers, a specialized class of classifiers whose boundaries can be adjusted by a scalar parameter. This adjustability facilitates the construction of safety regions that offer reliability in classification by adjusting boundaries to meet specific performance metrics, such as minimizing false positives. 

Following this, we integrate adjustable classifiers with two statistical approaches: probabilistic scaling and conformal prediction. These methods construct the desired safety regions with probabilistic assurances, which can be applied to machine learning classifiers without specific assumptions on the underlying data distribution.

\subsubsection{Adjustable Classifiers.}

Consider an input space $X \subseteq \mathbb{R}^d$ and an output space $Y = \{-1, +1\}$, where a binary adjustable classifier is defined as:

\begin{equation}
    \phi_\theta(x, \rho) = 
    \begin{cases}
      +1, & \text{if } f_\theta(x, \rho) < 0,\\
      -1, & \text{otherwise.}
    \end{cases}
\end{equation}
Here, $f_\theta : X \times \mathbb{R} \rightarrow \mathbb{R}$ is the classifier predictor. The function $f_\theta(x, \rho)$ is dependent on tunable hyperparameters $\theta$ and an adjustable scalar $\rho$, which shifts the classification boundary, allowing for control over the classifier’s performance, such as adjusting the false positive rate. Any classifier, $\hat{f}(x)$, can be made adjustable by adding $\rho$ as an offset: $f_\theta(x, \rho) = \hat{f}_\theta(x) + \rho$. A Support Vector Machine (SVM) classifier can thus be adapted as $f_\theta(x, \rho) = w^T \varphi(x) - b + \rho$, where $w$ is the vector of the learned weights, $\varphi$ is a feature map and $b$ is the offset.

For each point $x$, we define $\bar{\rho}(x)$ as the $\rho$ value where $x$ is on the decision boundary, i.e., $f_\theta(x, \bar{\rho}(x)) = 0$. This framework enables defining the $\rho$-safe set:

\begin{equation}
    S(\rho) = \{ x \in X : f_\theta(x, \rho) < 0 \},
\end{equation}
which is the region classified as safe (+1). However, this set alone does not ensure a probabilistic guarantee, only that the classifier predicts +1 within this region. To obtain a probabilistic safety region $S_\epsilon$ that satisfies $P(y = -1 | x \in S_\epsilon) \leq \epsilon$ with confidence $1 - \delta$, techniques from the field of order statistics can be used, such as probabilistic scaling and conformal prediction.

\subsubsection{Probabilistic Scaling.}

Probabilistic scaling (SP) constructs safety regions based on the order statistics of the calibration dataset $Z_c = \{(x_i,y_i)\}_{i=1}^{n^c} \subset X \times Y$. We define the generalized maximum $\text{max}_{(r)}(\Gamma)$ for a set $\Gamma = \{\gamma_i\}_{i=1}^n\in\mathbb{R}^n$ as the $r$-th largest value, ensuring no more than $r - 1$ elements of $\Gamma$ exceed $\text{max}_{(r)}(\Gamma)$. Assuming a continuous and monotonically increasing $f_\theta(x, \rho)$ on $\rho$, probabilistic scaling yields a $\rho_\epsilon$ such that:

\begin{equation}
    S_\epsilon = S(\rho_\epsilon),
\end{equation}
where $\rho_\epsilon$ is computed to satisfy $P(y = -1 | x \in S_\epsilon) \leq \epsilon$ with probability $1 - \delta$. Details and assumptions for this method are provided in \cite{carlevaro2023probabilistic}.

\subsubsection{Conformal Prediction.}

Conformal prediction (CP), as developed in \cite{vovk2015cross}, provides a post-hoc assessment of classification conformity. Using a score function $s : X \times Y \rightarrow \mathbb{R}$, such as $s(x, \hat{y}) = -\hat{y} \bar{\rho}(x)$, that encodes the agreement between a sample $x$ and a candidate label $\hat{y}$, CP defines a prediction region $C_\epsilon(x)$:

\begin{equation}
    C_\epsilon(x) = \{ \hat{y} \in \{-1, +1\} : s(x, \hat{y}) \leq s_\epsilon \},
\end{equation}
ensuring marginal coverage $P(\hat{y} \in C_\epsilon(x)) \geq 1 - \epsilon$. From this, the conformal safety region $\Sigma_\epsilon$ for input $x$ with class +1 can be derived, i.e.:
\begin{equation}
    \label{eq:conformal_safety_region}
    \Sigma_\varepsilon = \{x\in X\,:\, s(x,+1)\le s_\varepsilon, \ s(x,-1)>s_\varepsilon\},
\end{equation}
which is such that $S_\epsilon \subseteq \Sigma_\epsilon$. In \cite{carlevaro2024conformal} is shown that if $s_\epsilon \leq 0$ then $S_\epsilon = \Sigma_\epsilon$, achieving the desired confidence region. Details about conformal safety region (CSR) are provided in \cite{carlevaro2024conformal}.

In the following, we will denote the safety region $S_\varepsilon$ using the two methods as $S_\varepsilon^{PS}$ for the probabilistic scaling method and  $S_\varepsilon^{CP}$ for the conformal prediction method.

\subsection{Rule-Based Models Background}\label{sec:rulebasedback} 

Rule-based classifiers are machine learning models that provide outputs as sets of decision rules (rulesets), offering interpretability \cite{molnar2020interpretable,make6030101}. Different techniques for rule-based classification are typically grouped by their scope, based on whether they aim at providing explanations globally valid on the whole dataset, or locally on specific instances \cite{longo2024explainable}.

\subsubsection{Global Rule-Based Classifiers.}

In this kind of approach, the model learns a set of rules that represent the entire logic of the dataset, making it suitable for application to any data sample. The rules are “native” as they arise directly from the learning process without needing any intermediary steps. 

Formally, a rule-based classifier trained on a dataset $T = \{(x_j, y_j)\}_{j=1}^N \in X \times Y$, where $ x_j \in \mathbb{R}^d$ and $ y_j \in \{-1, +1\} $, generates a ruleset $ \mathcal{R} = \{r_k\}_{k=1}^M $. Each rule $ r_k $ has a premise, or antecedent, expressed as a conjunction of conditions:
\[
\textit{premise}(r_k) = \bigwedge_{i=1_k}^{N_k} c_{ik},
\]
where each condition $ c_{ik} $ specifies an interval on the input features, either bounded, left-bounded, or right-bounded. The rule’s consequence specifies the target class $ \hat{y}_k \in \{-1, +1\} $ associated with the premise.

\subsubsection{Local Rule Extraction via Anchors.} Anchors is a model-agnostic local rule extraction technique that generates high-precision rules for explaining individual predictions of any black-box classifier. While locally faithful, these rules also hold in a neighborhood (or perturbation space) of the instance being explained. An anchor $ A $ for an instance $ x $ is a set of predicates that satisfies the precision threshold $ \lambda_{\text{prec}} $ with a confidence level $ 1 - \delta $:
\[
\Pr\{ \text{Prec}(A) \geq \lambda_{\text{prec}} \} \geq 1 - \delta,
\]
where $ \delta \in [0, 1] $ and $ \lambda_{\text{prec}} \in [0, 1] $ sets the precision requirement for the anchor. Precision $ \text{Prec}(A) $ is defined as:
\[
\text{Prec}(A) = \mathbb{E}_{D_x(z|A)} \left[ \mathbbm{1}_{f(x) = f(z)} \right],
\]
where $ f $ is the black-box model, and $ D_x(z|A) $ is the distribution of perturbations $ z $ around $ x $ when the anchor applies. Optimal anchors are searched using reinforcement learning, and the process is formulated as a combinatorial optimization problem:
\[
\max_A \quad C(A) \quad \text{s.t.} \quad \Pr\{ \text{Prec}(A) \geq \lambda_{\text{prec}} \} \geq 1 - \delta,
\]
where $ C(A) $ represents the coverage for candidate anchor $ A $.

\subsubsection{Rule Evaluation.}
For rule-based classifiers, the performance of each rule $ r_k $ can be measured with two key metrics—coverage $ C(r_k) $ and error $ E(r_k) $—that allow us to evaluate the rule’s ability to generalize to unseen data. Before defining the metrics, it is worth underlining that a `positive' instance, in this context, refers to any instance that satisfies the considered rule, regardless of the output class. Similarly, the term `negative' is used to denote the case when points do not satisfy the rule.

\noindent \textbf{Coverage} $ C(r_k) $: this measures the proportion of correctly classified positive samples by rule $ r_k $:
\[
C(r_k) = \frac{\text{TP}(r_k)}{\text{TP}(r_k) + \text{FN}(r_k)},
\]
where $ \text{TP}(r_k) $ is the count of true positives and $ \text{FN}(r_k) $ is the count of false negatives associated with the rule.

\noindent \textbf{Error} $ E(r_k) $: it represents the proportion of false positives in the total predictions, helping to gauge rule precision:
\[
E(r_k) = \frac{\text{FP}(r_k)}{\text{TN}(r_k) + \text{FP}(r_k)},
\]
where $ \text{FP}(r_k) $ and $ \text{TN}(r_k) $ represent false positives and true negatives, respectively.

Combined, coverage and error contribute to the \textbf{relevance} $ R(r_k) $ of each rule, which indicates its generalizability:
\[
R(r_k) = C(r_k) \cdot (1 - E(r_k)).
\]

\subsection{Simulation of Social Robotics Navigation}\label{sec:navground}

Hinted by its name, Navground\footnote{\url{https://github.com/idsia-robotics/navground}} is a playground to experiment with navigation algorithms. The Navground social navigation simulator allows for experimentation with various navigation algorithms. At its core, the simulator operates with multi-agent (robots) systems that carry out specific navigation tasks, ensuring they avoid collisions with both static obstacles and other robots, and deadlocks. 
Each robot is represented as a circular disc, with its state defined by a $2D$ pose ($x$ and $y$), its orientation angle %the angle 
and its \emph{twist}; the velocity of the robot in the $x$-axis direction, its velocity in the $y$-axis direction and its rotational speed, in radians per second, around its central axis, which represents how fast the robot turns. Robots navigate using one of several reactive navigation behaviors, which consider the current state of the environment to generate control commands that guide them toward their targets while avoiding collisions. 

In this article, robots are modelled after the Thymio robot, a small robot with a size of 8 cm and two-wheel differential-drive kinematics, which is a very common kinematics\footnote{Kinematics is the study of the relationship between a robot's joint coordinates and its spatial layout, and is a fundamental and classical topic in robotics.} shared by many ground robots. 
Each simulated robot executes a specific behavior. 
In this article, robots will follow the \emph{human-like} (HL) navigation~\cite{HLikeJerome}, which is a bio-inspired, computationally light, local navigation algorithm for robotics, adapting a heuristic model for pedestrian motion. It addresses engineering aspects such as trajectory effectiveness and scalability, as well as societal aspects by producing human-friendly, predictable trajectories. In other words, this behavior is inspired by the way the pedestrian moves. 

The HL navigation algorithm operates in regular time intervals by following three key steps: first, it selects the best direction toward the target while avoiding potential collisions by considering a safety margin around the robot and the velocity of nearby entities. Second, it determines an appropriate speed that allows the robot to stop within a safe distance if needed. Finally, the velocity is smoothly adjusted over time to ensure natural movement transitions. The safety of the resulting trajectories depends on various parameters (see Tab. \ref{tab:parameters} for the explanation of some HL parameters), such as the safety margin $\sigma$, which helps account for modeling and perception errors.

\begin{table}[h!]
    \centering
    \caption{Description of HL Parameters}
    \label{tab:parameters}
    \begin{tabular}{|c|c|}
        \hline
        \textbf{Parameter} & \textbf{Description} \\
        \hline
        \hline
        $v_{\text{opt}}$ & The desired optimal speed \\
        $\tau_{\text{rot}}$ & The relaxation time to rotate towards a desired orientation \\
        $\sigma$ (safety margin) & The minimal safety margin (distance) to keep away from obstacles or other robots \\
        $\eta$ (eta)& The time that the behavior keeps away from collisions \\
        $\tau$ (tau)& The relaxation time controlling the smoothness of the motion \\ \hline
    \end{tabular}
\end{table}

\begin{figure}[h!]
    \centering
    \includegraphics[width=0.8\textwidth]{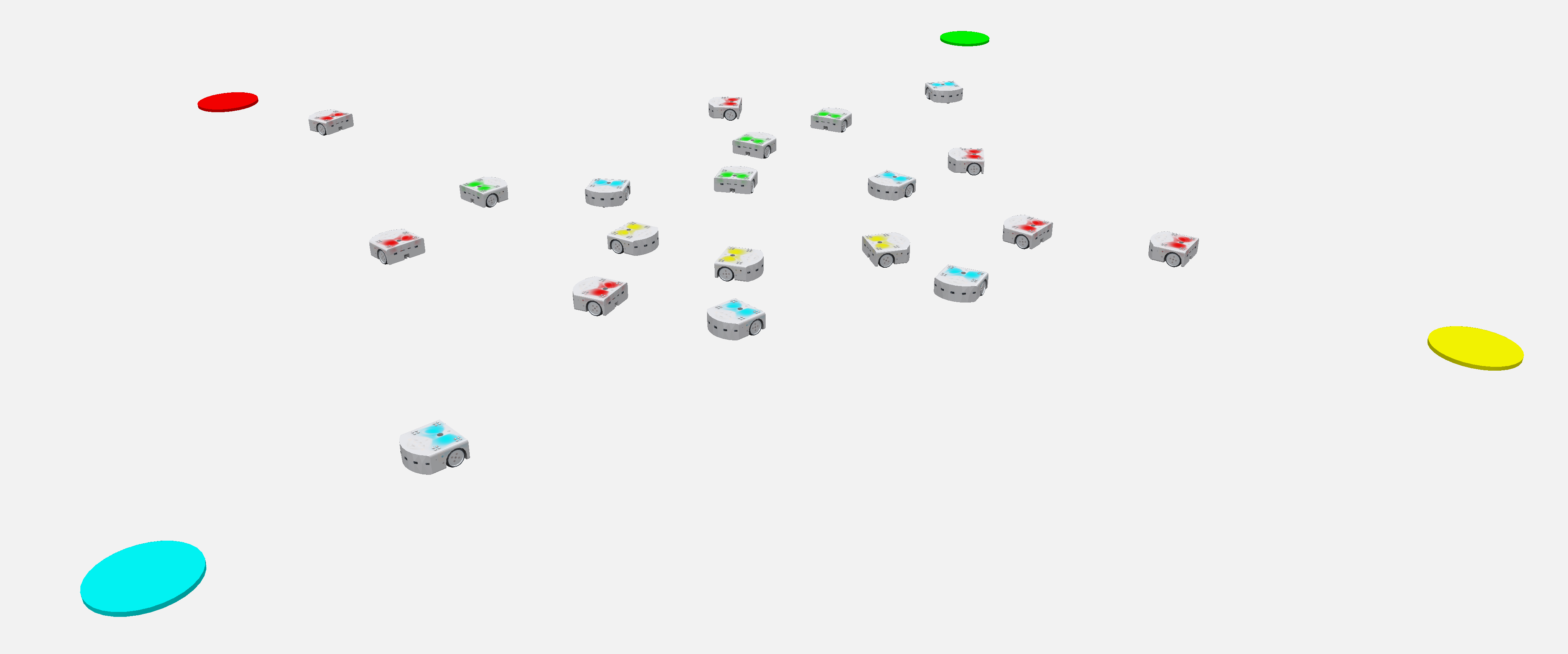}
    
    \caption{
    % Point cloud representation of robot positions in the \emph{crossings} scenario. In this setup, 
    In the \emph{crossings} scenario, simulated robots navigate between predefined waypoints (colored cylinders) along the x-axis and y-axis, creating intersecting flows at the center.
    } 
    \label{fig:crossScenario}
    
\end{figure}

Additionally, Navground provides different scenarios, where robots can navigate, in order to test and analyze their behavior. In this article, we will use the \emph{crossings} scenario. In this scenario, we define the variable $s$ as the length of each side of the square area containing the target waypoints and define the operational space for robots. The four target waypoints are located at the following coordinates:  $(\frac{-s}{2}, 0)$, $(\frac{s}{2}, 0)$, $(0, \frac{-s}{2})$, and $(0, \frac{-s}{2})$. Half of the robots are tasked to pendle between the two vertically aligned waypoints, and half between the horizontally aligned waypoints (see Fig. \ref{fig:crossScenario}). The scenario tests how robots cross in the middle, where the 4 opposing flows meet. In the case of the cross scenario, we must be clear that the positions of the robots go on the $x$-axis from $\frac{-s}{2}$ to the value set in $\frac{s}{2}$ of the scenario, and on the $y$-axis similarly.

In addition, in the context of social navigation between robots, deadlocks and collisions refer to two different types of issues that can arise during the interaction and navigation of multiple robots in a shared environment. A deadlock in robot navigation occurs when one or more robots become trapped in a situation where they cannot move due to the positions of other robots, waiting for the other to move in order to take up the movement again. A collision occurs when two or more robots collide or when a robot collides with static obstacles in the environment. This can result from uncoordinated movements or errors in navigation planning and control. We will also focus on these two negative events in experiments. 
Notice that a simulation can be considered safe if there is no collision between robots, efficient if no robot enters a deadlock state at any point during the simulation, and compliant if there are neither collisions nor deadlocks.

\section{Topology-Driven Safety Methodology}\label{sec:methodology}

Once all preliminary concepts have been defined, we can present the core contribution of this paper: a \textbf{topology-driven methodology} for defining safety regions in robot simulations. By combining topological tools with entropy-based analysis, we provide a systematic approach to avoid/prevent unsafe events such as robot collisions or deadlocks. 

At a high level, our methodology consists of the following steps (see Fig.~\ref{fig:methodologyillustrative} for an example):

\begin{figure}
    \centering
    \includegraphics[width=\textwidth]{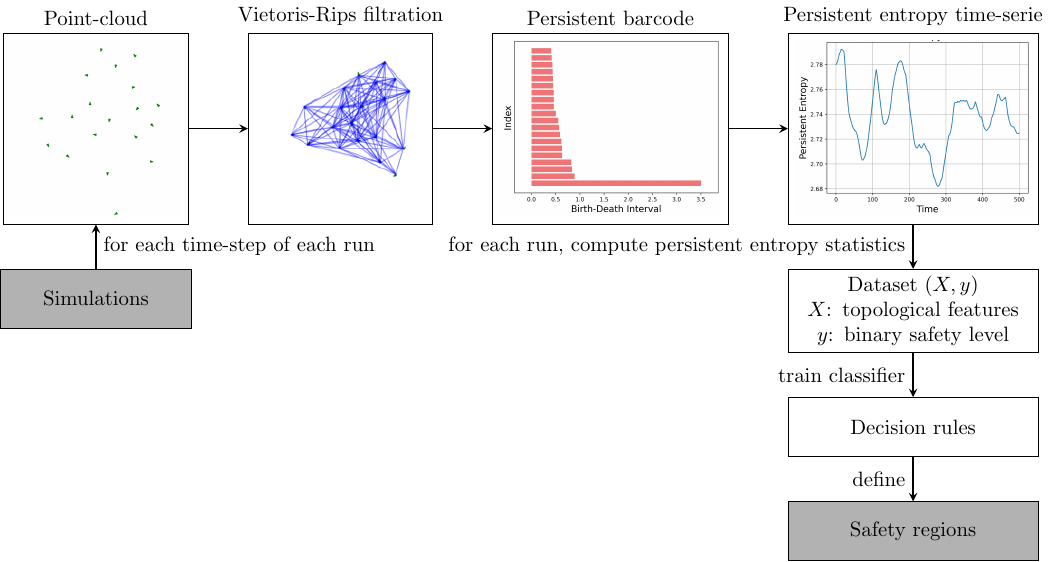}
    \caption{Illustrative example of methodology pipeline.}
    \label{fig:methodologyillustrative}
\end{figure}

\begin{enumerate}
    \item \textbf{Simulate:}  Generate multiple navigation scenarios where a fleet of mobile robots moves between opposing targets, potentially leading to collisions or deadlocks.
    \item \textbf{Build a dataset:} Extract topological features from each simulation and assign a binary safety label based on observed events (e.g., presence or absence of collisions). This process consists of the following steps:

     \begin{enumerate}
     \item \textbf{Extract point clouds:} From each  simulation run performed by Navground, we extract the positions of all robots at each time step, resulting in $n$ point clouds. Then, for each point cloud, we perform steps b--e.
     \item \textbf{Compute persistent homology:} We compute the persistent homology of the point cloud, using the Vietoris-Rips complex. This yields a \textbf{persistence barcode}, focusing only on 0-dimensional features.
     \item \textbf{Analyze birth-death intervals:} From the persistence barcode, we extract the lengths of the intervals that correspond to the 0-dimensional topological features.
     \item \textbf{Normalize interval lengths:} We compute the sum of all interval lengths and normalize each length by this sum, obtaining a probability distribution of interval lengths of the topological features.
     \item \textbf{Calculate persistent entropy:} We compute the entropy of the topological features probability distribution, using the Shannon entropy formula Eq.~\eqref{eq:entropy}. 
     See Fig. \ref{fig:methodologyEntropy} for an example of a point cloud corresponding to the position of the robots in a specific time step, the corresponding persistence barcode of dimension 0, and the corresponding persistent entropy.
     \item \textbf{Generate persistent entropy time 
     series and compute statistical parameters}. Once the above process is done for each time step of the simulation, we obtain a persistent entropy time series for the whole simulation. Finally, we compute four key statistical parameters for this time series: \textbf{mean,  median, standard deviation and interquartile range (IQR)}. See Fig. \ref{fig:entropyTS} for an example of persistent entropy time series over a 500-step simulation.
 \end{enumerate}

    Once the topological features are extracted, we construct a labeled dataset to train a classifier. The dataset is defined as: 

    \begin{itemize}
    \item Input $x$: A feature vector extracted from the topological data of the simulation, defined as: $x = (meanEntropy, medianEntropy, stdEntropy, iqrEntropy)$. These features are derived from the 6-step methodology explained above.
    \item Output $y$: A binary label indicating whether the simulation is safe(+1) or unsafe(-1). The definition of ``safe'' depends on the context. Given a set of simulations, we assign each simulation a binary level indicating its safety level, depending on the avoidance event, for example, for safe analysis (collision or not collision): \[
    y_i =
    \begin{cases} 
    +1 & \text{if the number of collisions during the simulation $i$ equals $0$,}
    %= 0, 
    \\ 
    -1 & \text{if the number of collisions during the simulation $i$ is greater than $0$}.
    \end{cases}.
    \]

\end{itemize}   

\begin{figure}[h!]
\centering
\begin{subfigure}{0.48\textwidth}
    \includegraphics[width=\textwidth,height=\textwidth]{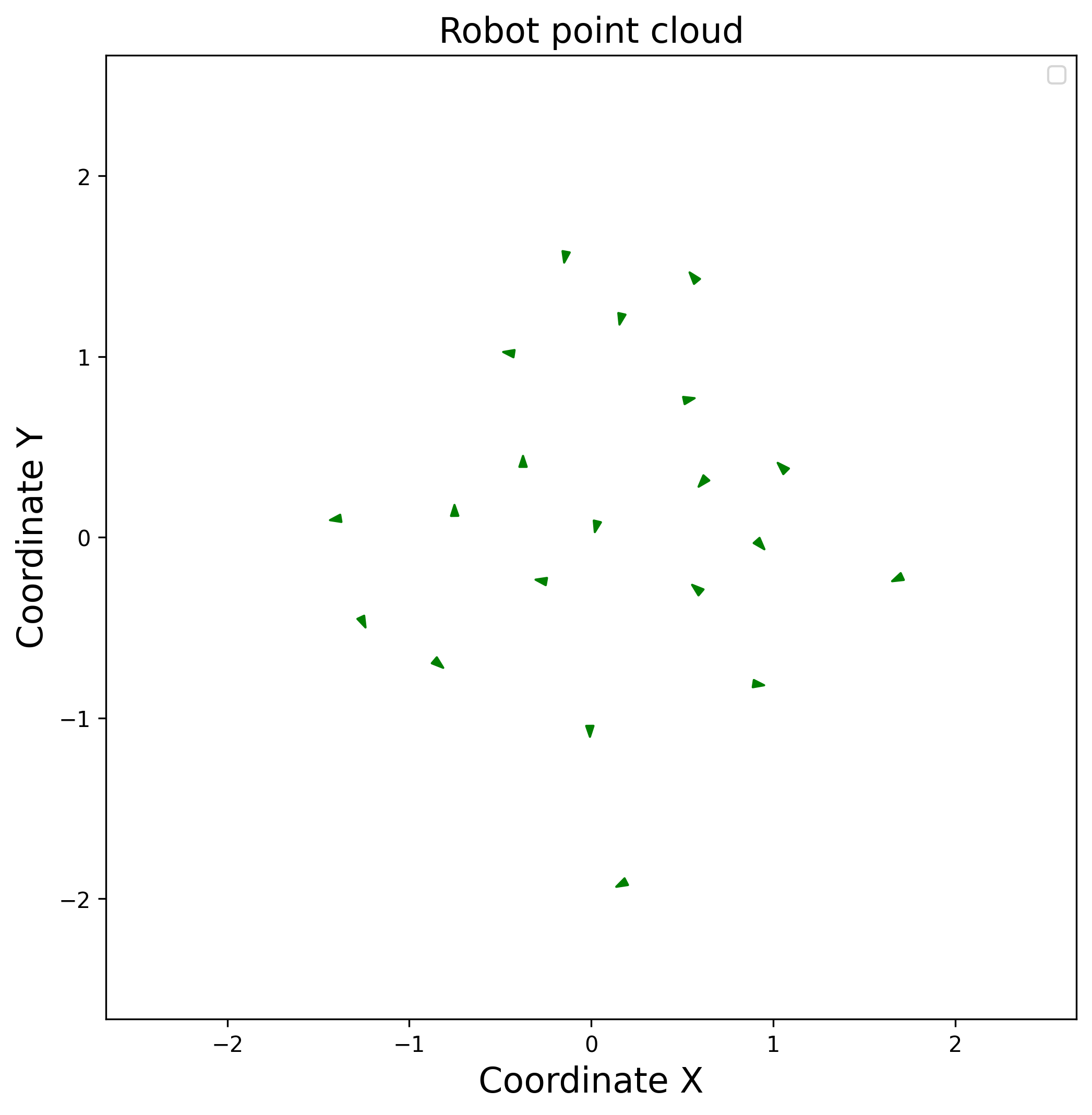}
    \caption{Point cloud example of the robots in a specific time step.}
\end{subfigure}
\hfill
\begin{subfigure}{0.48\textwidth}
    \includegraphics[width=\textwidth,height=\textwidth]{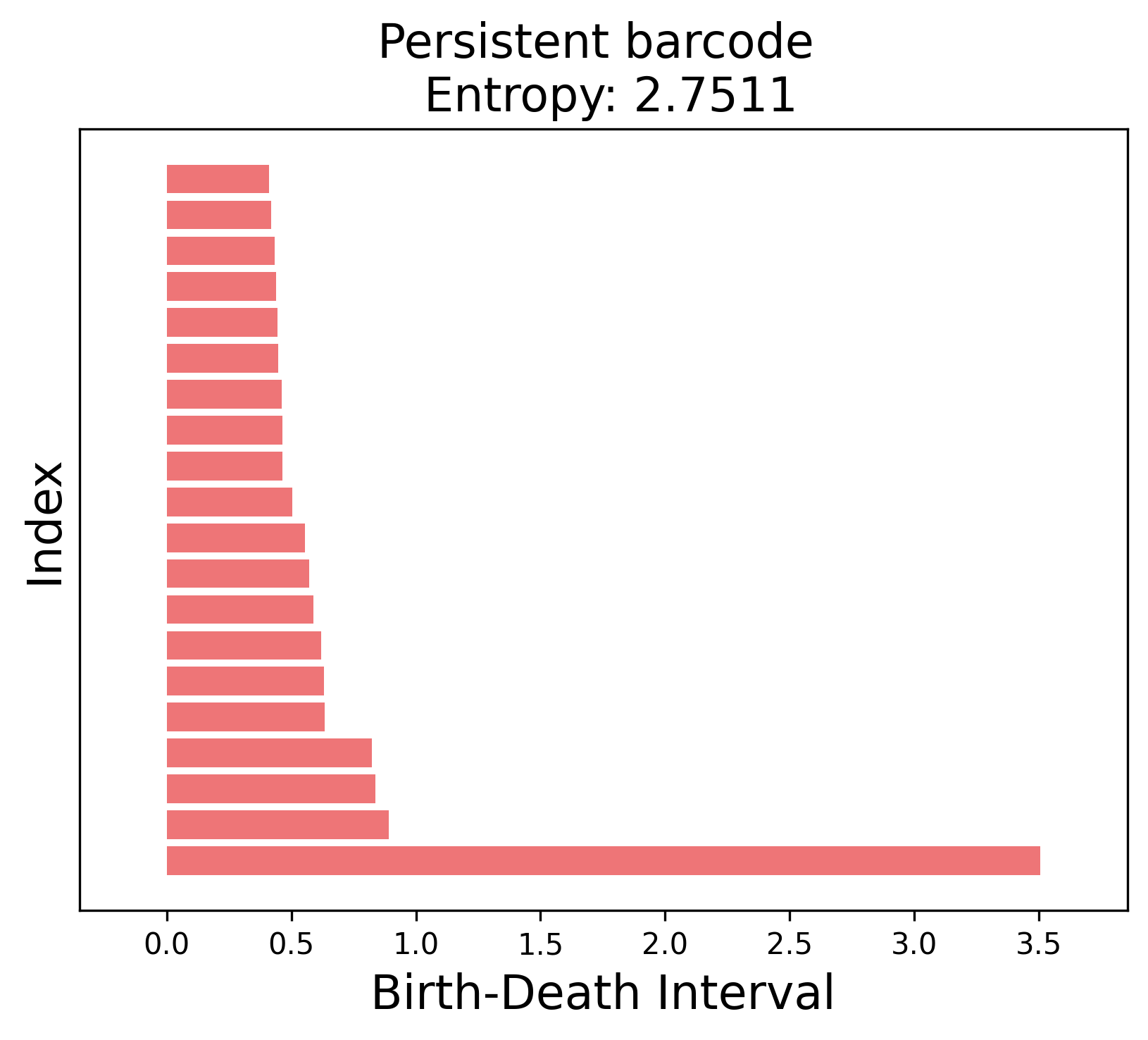}
    \caption{Persistence barcode of dimension 0 with corresponding persistent entropy
    of $2.7511$.}
\end{subfigure}
\caption{Illustrative example of computing the persistent entropy of a robot point cloud.}
\label{fig:methodologyEntropy}
\end{figure}

\begin{figure}[h!]
    \centering
    \includegraphics[width=0.6\textwidth,height=0.4\textwidth]{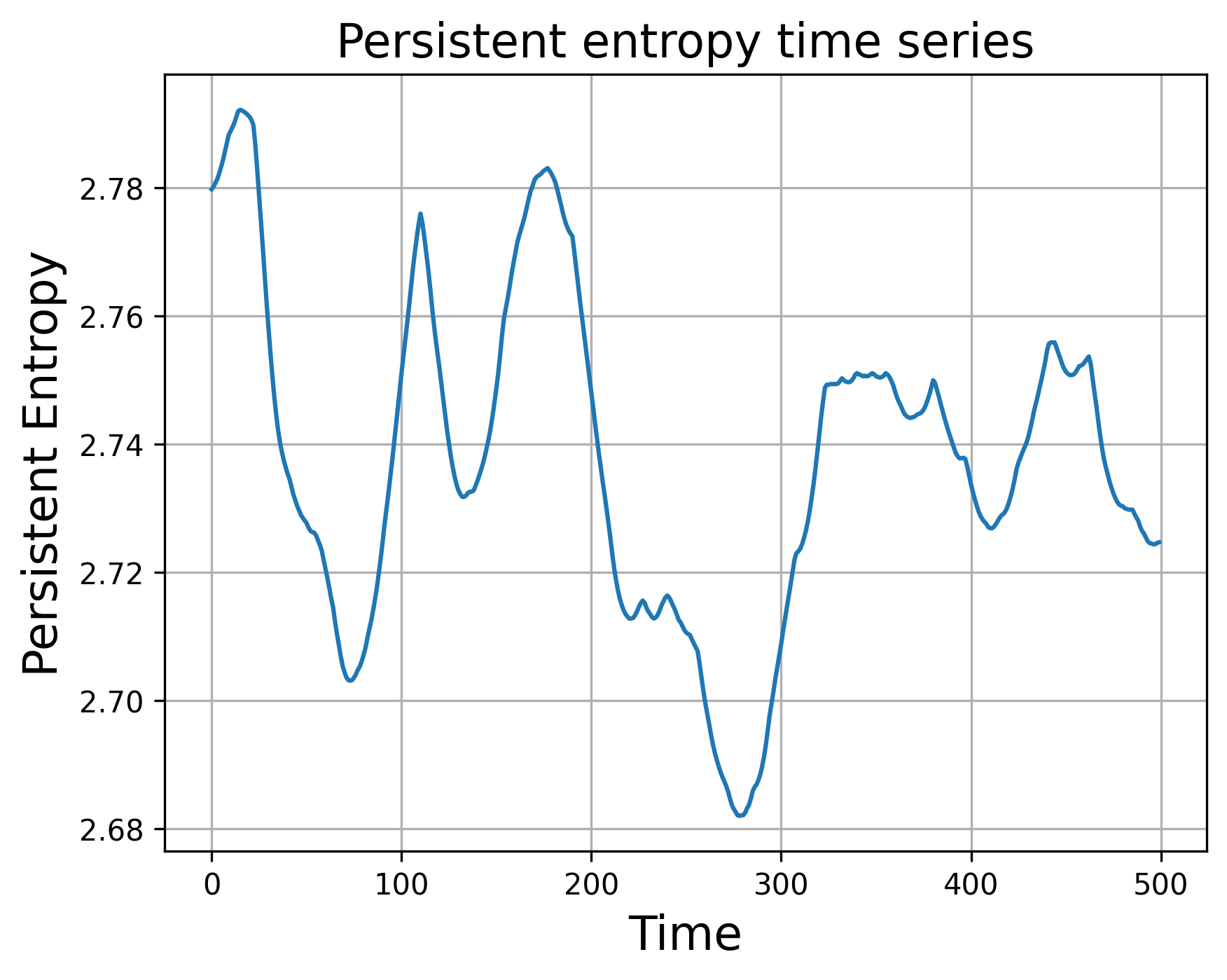}
    \caption{Persistent entropy time series example of a simulation.} \label{fig:entropyTS}
\end{figure}
    
    \item \textbf{Train a classifier:} Using this dataset, we train a Support Vector Machine (SVM) classifier with adjustable margins to learn the decision function separating safe and unsafe simulations.
    \item \textbf{Define safety regions:} Once the classifier is trained, we use it to construct probabilistic safety regions, which define zones in the topological feature space where simulations are likely to be safe. These regions are constructed using order statistics methods to the classifier, such as probabilistic scaling and conformal prediction. In addition, we extract local rules from these safety regions using Anchors.
\end{enumerate}

\begin{figure}[h!]
    \centering
    \includegraphics[width=\textwidth,height=0.4\textwidth]{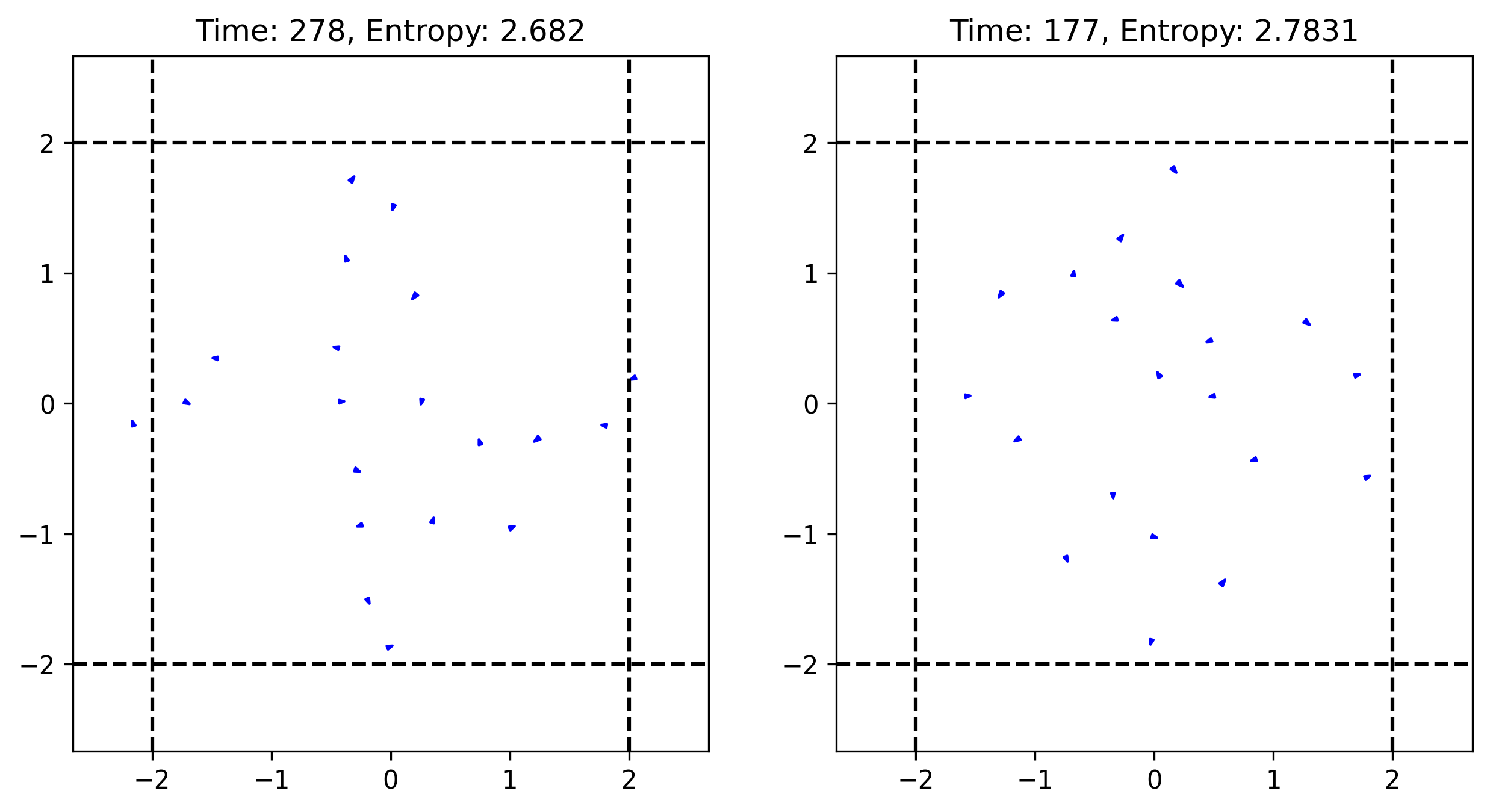}
    \caption{Point clouds of two different time steps with the corresponding persistent entropy on top. 
    } \label{fig:LowerMaxEntropyExample}
\end{figure}

Note that a specific time step in a simulation with a higher persistent entropy indicates a greater dispersion among the robots in relation how they are distributed around the space, without clusters of robots in certain areas and without empty areas or areas with few robots, being less likely to have empty areas and to form cluster of robots very grouped and another ones very far away, while a lower entropy indicates that less dispersion between the robots, being more likely to occupy specific regions of space and form clusters. See Fig. \ref{fig:LowerMaxEntropyExample} for an illustrative example where we can compare the robot point cloud from two different time steps. The point cloud on the left has lower entropy than the one on the right (where the points are more dispersed).

\section{Experiments}\label{sec:experiments}

In this section, we present the experiments and results we carried out by putting together the simulator, the safety regions and the illustrated topological techniques. After data collection through Navground simulator, we study the problem of collision avoidance, comparing our methodology with the one that uses behavior parameters ($\sigma, \eta \text{ and } \tau$, which are defined in Tab. \ref{tab:parameters}  for the construction of safety regions \cite{10.1007/978-3-031-63803-9_22} (instead of the topological features), then studying the problem of deadlock avoidance, and finally combining them for safe and efficient (free of collisions and deadlocks) navigations.

\subsection{Data Collection}\label{subsec:datacollection}

Using Navground, we generated a suitable dataset to study the safety of robots' movement while avoiding collisions, deadlocks, and both of them, via probabilistic scaling and conformal prediction methods, with interpretation via rule-based classifiers.
We executed $N = 10000$ simulation runs, each with 2000 time steps, each one separated from the previous one in 0.1 seconds, with a group of 20 robots modelled after the Thymio robot. Each simulated robot executes the HL navigation with the following parameters (behavior parameters of the simulations):

\begin{itemize}
    \item $v_{\text{opt}} = 0.12$ m/s
    \item $\tau_{\text{rot}} = 0.5$ s
    \item $\sigma$ sampled uniformly from $[0.0 \text{ m}, 0.1  \text{ m}]$ for collision avoidance, and from $[0.0 \text{ m}, 0.5  \text{ m}]$ for deadlock avoidance and for safe and efficient simulations (since there are hardly any simulations with deadlocks if we limit this parameter to 0.1).
    \item $\tau$ and $\eta$ sampled uniformly from $[0.0 \text{ s}, 1.0 \text{ s}]$
\end{itemize}

These HL parameters ($\sigma, \eta \text{ and } \tau$) have been sampled to generate simulations of all types, including safe, aggressive, cautious, or efficient simulations. Please, see Table~\ref{tab:parameters} for an explanation of the HL parameters.

For each simulation run, we apply the methodology explained previously (Section \ref{sec:methodology}), recording the following values:
\[
\textbf{x} = (meanEntropy, medianEntropy, stdEntropy, iqrEntropy).
\]

According to the computational cost of calculating persistent entropy, for instance, the calculation time for persistent entropy for a point cloud at a given time $i$ is less than 10 ms on a single core of a modern CPU (approximately 0.003 seconds), and the total time for computing the entire persistent entropy time series for a simulation with the mentioned characteristics is approximately 0.2 seconds. These times suggest that the method is efficient enough for real-time performance in practical applications.

Then, we assigned a binary label $y$ to each simulation through the following criteria, depending on the avoidance event:

\begin{itemize}
    \item Safe simulation (avoiding collision): \[
y =
\begin{cases} 
+1 & \text{if number of collisions} = 0, \\ 
-1 & \text{if number of collisions} > 0 
\end{cases}.
\]

    \item Efficient simulation (avoiding deadlocks): \[
y =
\begin{cases} 
+1 & \text{if number of deadlocks} = 0, \\ 
-1 & \text{if number of deadlocks} > 0 
\end{cases}.
\]

    \item Compliant simulation (avoiding both, collision and deadlocks): \[
y =
\begin{cases} 
+1 & \text{if number of collisions/deadlocks} = 0, \\ 
-1 & \text{if number of collisions/deadlocks} > 0 
\end{cases}.
\]

\end{itemize}

Finally, we obtained the dataset $T_{\text{nav}} = \{(\textbf{x}_j , y_j) \mid j = 1, \ldots, N\}$. In Section~\ref{sec:results}, we present and analyze the results obtained using reliable AI techniques for this dataset.

\subsection{Data Exploration}\label{sec.dataexploration}

\begin{figure}[h!]
    \centering
    \begin{subfigure}[b]{0.32\textwidth}
        \includegraphics[width=\textwidth,height=1.2\textwidth]{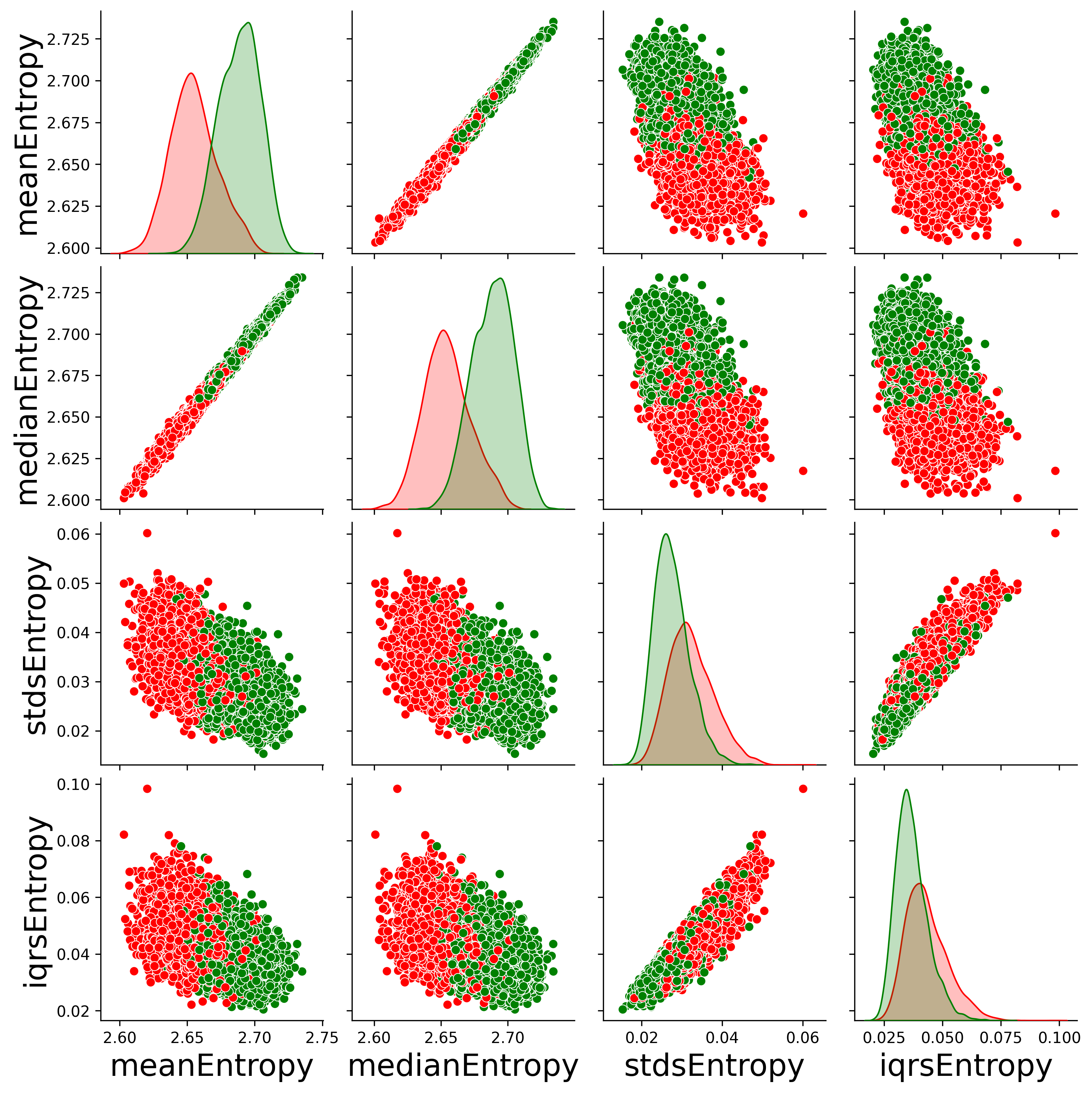}
        \caption{Collision: Green = ``no collision'', Red = ``collision''.}
        \label{fig:exploratoryCollision}
    \end{subfigure}
    \hfill
    \begin{subfigure}[b]{0.32\textwidth}
        \includegraphics[width=\textwidth,height=1.2\textwidth]{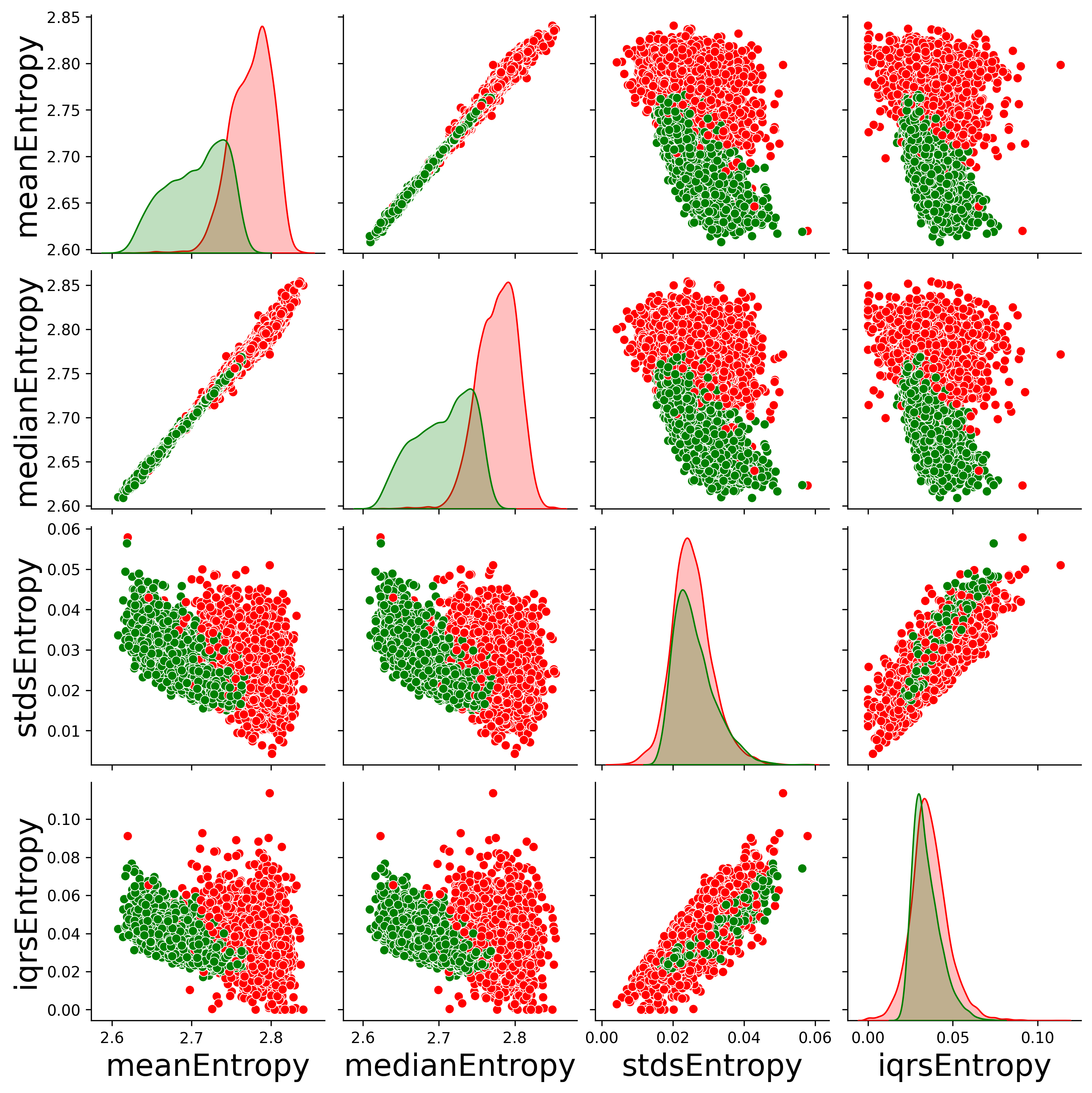}
        \caption{Deadlock: Green = ``no deadlock'', Red = ``deadlock''.}
        \label{fig:exploratorydeadlock}
    \end{subfigure}
    \hfill
    \begin{subfigure}[b]{0.32\textwidth}
        \includegraphics[width=\textwidth,height=1.2\textwidth]{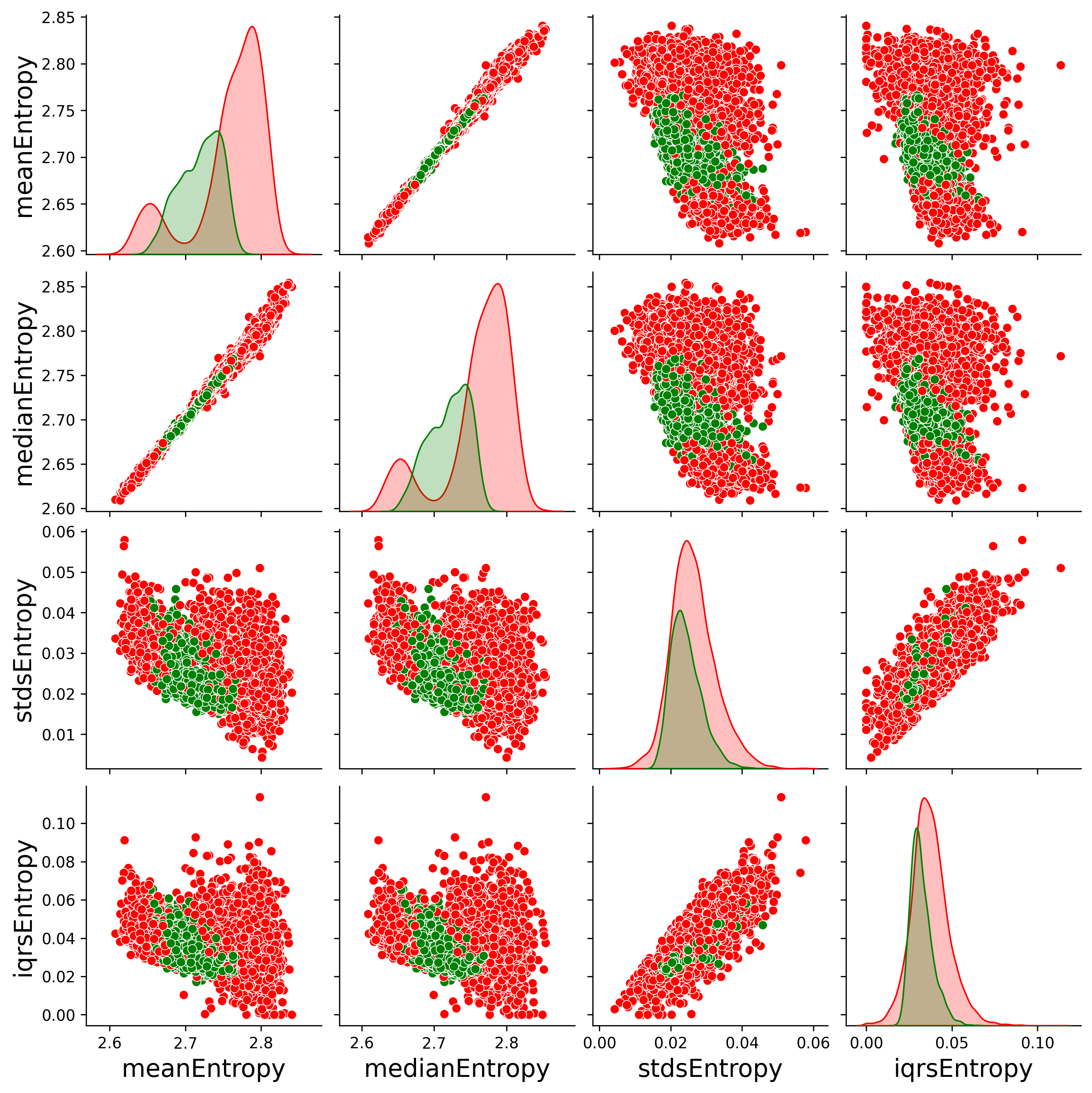}
        \caption{Safety: Green = ``safe'', Red = ``non safe''.}
        \label{fig:exploratorySafe}
    \end{subfigure}
    \caption{Pairwise class distributions of the features in $T_{\text{nav}}$.}
    \label{fig:exploratoryCombined}
\end{figure}

Before entering in the results in terms of confidence regions and explainability, a first visual inspection of how classes are distributed in the dataset is useful to understand the non-trivial nature of the problem. In Figs. \ref{fig:exploratoryCollision}, \ref{fig:exploratorydeadlock} and  \ref{fig:exploratorySafe}, the relationships among the four persistent entropy-based metrics — mean entropy, median entropy, standard deviation of entropy (stdEntropy), and interquartile range of entropy (iqrsEntropy) — are visualized for three binary classifications: collision avoidance, deadlock avoidance, and compliant simulations (collision and deadlock avoidance). Each plot displays pairwise scatterplots, overlaid with KDE (Kernel Density Estimation) curves along the diagonals to depict distributions for each class.

Across all three classifications — collision avoidance, deadlock avoidance, and compliant simulations (collision and deadlock avoidance) — \textbf{mean entropy and median entropy} stand out as the most informative features for distinguishing between classes. While stdEntropy and iqrsEntropy provide less individual discriminatory power, they may still offer complementary information when used in combination with the other metrics. These insights highlight the potential of persistent entropy-based metrics for identifying critical system states and suggest a strong basis for further model development and optimization.

Lower values of mean entropy and median entropy are more closely associated with the ``collision'' class, indicating that decreased entropy in the system could signal potential collisions between robots. Higher values of mean entropy and median entropy are more closely associated with the ``deadlock'' class, indicating that increased entropy in the system could signal potential deadlock situations. In this way, we can observe how safe and efficient safe simulations, without collisions or deadlocks, have relatively stable values for mean entropy and median entropy, unless an extreme value occurs, either low or high.

The goal of this analysis is to characterize such a region in an interpretable way, via post-hoc rule extraction from confidence
regions.

\subsection{Results}\label{sec:results}

First, we will compare training a rule-based model with the behavior parameters (as is done in \cite{10.1007/978-3-031-63803-9_22}) against by training it with the topological features that have been extracted using the methodology proposed in this paper for the avoidance of collisions. For that, we will use Skope Rules. Skope Rules\footnote{https://github.com/scikit-learn-contrib/skope-rules} is a rule-based ML algorithm, that learns interpretable and diversified rules for “scoping” a target class
of interest, i.e. detecting samples from this class with high precision. In practical applications, skope-rules model has been applied to several safety-critical classification tasks, as well as anomaly detection problems or cluster description.

\begin{table}[ht]
\centering
\caption{Performance comparison between the adopted rule-based models. The first
column reports the number of generated rules. The other columns refer to the following
metrics (expressed in \%): accuracy (ACC), F1-score (F1), true positive rate (TPR),
false positive rate (FPR), false negative rate (FNR), true negative rate (TNR).}
\begin{tabular}{@{}lccccccc@{}}
\toprule
 & \# of rules & ACC   & F1    & TPR   & FPR   & FNR   & TNR   \\ \midrule
SkopeRules-Behavior Parameters  & \pgfmathparse{4+5}\pgfmathprintnumber[precision=0]{\pgfmathresult} & \pgfmathparse{(0.738+0.835)/2*100}\pgfmathprintnumber[precision=1]{\pgfmathresult}  & \pgfmathparse{(0.6507+0.8515)/2*100}\pgfmathprintnumber[precision=1]{\pgfmathresult}  & \pgfmathparse{(0.5105+0.9061)/2*100}\pgfmathprintnumber[precision=1]{\pgfmathresult}  & \pgfmathparse{100-(0.9464+0.7532)/2*100}\pgfmathprintnumber[precision=1]{\pgfmathresult}  & \pgfmathparse{100-(0.5105+0.9061)/2*100}\pgfmathprintnumber[precision=1]{\pgfmathresult}  & \pgfmathparse{(0.9464+0.7532)/2*100}\pgfmathprintnumber[precision=1]{\pgfmathresult} \\ \midrule
SkopeRules-Topological Features & \pgfmathparse{11+9}\pgfmathprintnumber[precision=0]{\pgfmathresult} & \pgfmathparse{(0.837+0.855)/2*100}\pgfmathprintnumber[precision=1]{\pgfmathresult}  & \pgfmathparse{(0.8137+0.8776)/2*100}\pgfmathprintnumber[precision=1]{\pgfmathresult}  & \pgfmathparse{(0.8416+0.9012)/2*100}\pgfmathprintnumber[precision=1]{\pgfmathresult}  & \pgfmathparse{100-(0.8336+0.7919)/2*100}\pgfmathprintnumber[precision=1]{\pgfmathresult}  & \pgfmathparse{100-(0.8416+0.9012)/2*100}\pgfmathprintnumber[precision=1]{\pgfmathresult}  & \pgfmathparse{(0.8336+0.7919)/2*100}\pgfmathprintnumber[precision=1]{\pgfmathresult}  \\ \bottomrule
\end{tabular}
\label{tab:comparisonSkrules}
\end{table}

Table \ref{tab:comparisonSkrules} highlights the superior performance of SkopeRules using topological features compared to behavior parameters. For a proper interpretation of the results, we remark that, based on the labeling criteria defined in Sec. \ref{subsec:datacollection}, the positive class (i.e., $y=+1$) refers to our target situation (that is, either absence of collisions, deadlocks, or absence of both). As a result, true positives here denote their correct classification, with false positives reflecting how many hazardous cases (presence of collisions, deadlocks, or both, labeled as $y=-1$) are missed. In contrast, true negatives are associated to the dangerous situations being correctly classified, with false negatives representing false alarm cases.

The model using topological features as input features achieves higher accuracy (84.6\% vs. 78.6\%), F1-score (84.6\% vs. 75.1\%), and true positive rate (87.1\% vs. 70.8\%) while significantly reducing the false negative rate (12.9\% vs. 29.2\%). Although its true negative rate (81.3\% vs. 85\%) is slightly lower, and the false positive rate (18.7\% vs. 15\%) is higher, these trade-offs are outweighed by its improved balance across metrics. Remember that the positive class is the non-collision class (+1).

The use of 20 rules, compared to 9 for behavior parameters, suggests that the model that using topological features as input features, captures more complex relationships, enhancing its robustness. Models with fewer rules have the advantage of being more interpretable, but the richest ones may generate more fine-grained rules with better discriminative ability. Overall, topological features provide a more discriminative representation, making the model using them as input features preferable for scenarios where accuracy and minimising false positives are prioritised.

Also, we will compare the safety regions generated using these two types of input features. For that, the techniques detailed in Section \ref{sec:confregback} to derive the safety regions using Conformal Prediction and Probabilistic Scaling were applied by adopting an SVM as the classifier. Specifically, we considered a Gaussian kernel SVM, with a sigma set to 0.5, with regularization parameter set to 0.3, and a weighting of 0.5. These hyperparameters were selected manually to ensure a fair and controlled comparison across the different types of input features, rather than focusing on obtaining the best possible SVM performance. With this base model, prior to any error control, we achieve the following performance: 78.2\% accuracy, 80.6\% F1 Score, 82.8\% TPR, 72.7\% TNR, 27.3\% FPR, 17.2\% FNR using behavior parameters as input and 85\% accuracy, 88.1\% F1 Score, 91.6\% TPR, 75\% TNR, 25\% FPR, 8.4\% FNR using topological features as input.

\begin{table}[h!]
    \centering
    \caption{Performance comparison between the adopted techniques for finding safety regions at $\epsilon = 0.1$. The first column refers to the input features used for training the classification models, the second refers to the adopted techniques for finding safety regions, and the third one reports the optimal scaling parameter. The other columns refer to the following metrics (expressed in \%): accuracy (ACC), F1-score (F1), true positive rate (TPR), false positive rate (FPR), false negative rate (FNR), true negative rate (TNR). Probabilistic Scaling denoted as PS, and Conformal Prediction denoted as CP.}
    \begin{tabular}{@{}lcccccccc@{}}
        \toprule
         & Method & $\rho_\epsilon$ & ACC & F1 & TPR & FPR & FNR & TNR \\
        \midrule
        \multirow{2}{*}{Behavior parameters} & PS & 0.48 & 77.9 & 77 & 67.8 & 9.9 & 32.1 & 90.1 \\
        & CP & 0.44 & 78.5 & 77.8 & 69 & 10.1 & 31 & 89.9 \\
        \midrule
        \multirow{2}{*}{Topological Features} & PS & 0.42 & 79.8 & 81.6 & 74.3 & 11.9 & 25.7 & 88.1 \\
        & CP & 0.21 & 83.7 & 86.2 & 84.1 & 16.9 & 15.9 & 83.1 \\
        \bottomrule
    \end{tabular}
    \label{tab:safetyregionsComparison}
\end{table}

Topological features consistently show better performance than behavior parameters across all metrics, both in baseline classification and in safety region generation. Among the methods for generating safety regions (see Table \ref{tab:safetyregionsComparison}), Conformal Prediction demonstrates superior accuracy and F1 scores, particularly with topological features, making it the preferred approach for ensuring robust safety guarantees.

Now, we want to derive interpretable approximations for both regions in the form of decision rules. Local rule extraction via Anchors was performed on a set of instances labeled as +1 by the adjustable classifiers and sampled at a small distance  $d \leq 0.05$ from their border.

\begin{table}[h!]
    \centering
    \caption{Performance comparison between the rules obtained using Anchors from the obtained adjustable classifiers (SVM). The first column refers to the input features used for training the adjustable classifiers, the second refers to the adopted techniques for finding safety regions, and the third column reports the number of generated rules (anchors). Covering and error percentages are reported for the logical union of the anchors being tested with respect to the labels assigned via PS ($S_\varepsilon^{PS} output$) and CP ($S_\varepsilon^{CP} output$) (Method labels column), and the real labels (Ground Truth column). Probabilistic Scaling is denoted as PS, and Conformal Prediction is denoted as CP.}
    \begin{tabular}{@{}l|c|c|cc|cc@{}}
        \toprule
         &  &  & \multicolumn{2}{c}{Method labels} & \multicolumn{2}{c}{Ground Truth} \\ & Method & \# of anchors & Coverage & Error & Coverage & Error\\
        \midrule
        \multirow{2}{*}{Behavior Parameters} & PS & 5 & 86 & 36 & 73 & 29\\
        & CP & 5 & 85 & 35 & 73 & 29 \\
        \midrule
        \multirow{2}{*}{Topological Features} & PS & 2 & 79 & 0 & 69 & 9  \\
        & CP & 2 & 100 & 30  & 100 & 26\\
        \bottomrule
    \end{tabular}
    \label{tab:anchorsComparison}
\end{table}

The results in Table \ref{tab:anchorsComparison} clearly demonstrate the advantages of using topological features over behavior parameters in the context of local rule extraction with Anchors. Notably, the number of generated rules (anchors) is significantly reduced when using topological features—only 2 rules ($meanEntropy  > 2.68$ and $medianEntropy > 2.68$ for PS method, and $meanEntropy  > 2.66$ and $medianEntropy > 2.65$ for CP method. Entropy above these values refers to simulations free of collision situations.) compared to 5 rules for behavior parameters (the same 5 rules for both methods: $\eta > 0.75, \tau \leq 0.5, \tau \leq 0.25, \tau \leq 0.76$ and $\sigma > 0.08$). This reduction highlights a key advantage: the rules obtained with topological features are simpler, more concise, and easier to interpret, which enhances their applicability in decision-making tasks.

In addition to reducing the complexity of the rules, the performance also improves. For topological features, the error rates are markedly lower. For instance, under Probabilistic Scaling (PS), the Coverage Error drops to 9\%, compared to 29\% with behavior parameters. Even with Conformal Prediction (CP), the Coverage Error decreases to 26\% while achieving a perfect 100\% coverage. Despite having fewer rules, topological features maintain competitive or superior coverage compared to behavior parameters.

These findings emphasize that topological features not only simplify the decision rules by reducing their number but also improve their accuracy and reliability, making them a more effective and interpretable choice for defining local decision regions.

Once we have verified that the proposed methodology achieves better results for differentiating between safe (without robots collisions) and non-safe simulations, we proceed to extend its use to the case of distinguishing between efficient (without deadlocks) and non-efficient simulations. We follow exactly the same steps, but this time we do not compare the results/performance with those obtained using behavior parameters (SafetyMargin, Eta and Tau).

Using the SkopeRules method and topological features to classify simulations with deadlock cases versus efficient simulations without them, we derive a total of 18 rules, achieving highly positive performance: 88.6\% accuracy, 89\% F1-score, 93\% TPR, and 84\% TNR. This demonstrates that these features achieve even better performance for deadlock avoidance than for collision avoidance when using SkopeRules.

Next, we generate the safety regions using these features in the same manner as before. We again considered a Gaussian kernel SVM, with a sigma set to 0.5, a regularization parameter set to 0.3, and a weighting of 0.5. With this base model, prior to any error control, we achieve the following performance: 86\% accuracy, 82.3\% F1 score, 74.7\% TPR, 94.7\% TNR. Using PS method we achieve the following performance: $\rho_\epsilon$ equal to -0.12, 87.2\% accuracy, 84.6\% F1 score, 80.1\% TPR, 92.7\% TNR; and using the CP method we achieve the following performance: $\rho_\epsilon$ equal to 0.17, 83.6\% accuracy, 77.9\% F1 score, 66\% TPR, 97.3\% TNR.

Again, local rule extraction via Anchors was performed on a set of instances labeled as +1 by the adjustable classifiers and sampled at a small distance $d \leq 0.05$ from their border. The number of generated rules (anchors) is 2 (same for both methods: $meanEntropy \leq 2.75$ and $medianEntropy \leq 2.75$. Entropy below these values refers to simulations free of deadlock situations.), so again, the rules obtained are very simpler and easier to interpret. We evaluate the logical union of the anchors being tested with respect to the labels assigned via the method used, and the real labels, obtaining the following results:

- 91\% coverage and 21\% error for ground truth labels, and 100\% coverage and 22.5\% for method labels for PS method.

- 91\% coverage and 21\% error for ground truth labels, and 100\% coverage and 40\% for method labels for CP method.

We can see how with the PS method we obtain a better performance in terms of error, although the rules are exactly the same, obviously obtaining the same performance for the real labels. Considering the non-trivial nature of the problem and of approximating a complex SVM shape via hyper-rectangular shapes (rules) while keeping the error bound as low as possible, we can consider our results as a promising compromise between safety and transparency.

Thus, if we focus on distinguishing between safe and efficient (compliant) simulations (free of collisions and deadlocks) and not compliant simulations, we 
obtain very simple rules for both methods:

- $2.68 \leq meanEntropy \leq 2.75$ and $2.68 \leq medianEntropy \leq 2.75$ for PS method.

- $2.66 \leq meanEntropy \leq 2.75$ and $2.65 \leq medianEntropy \leq 2.75$ for CP method.

In summary, our results highlight the advantages of using topological features in rule-based explanations and the generation of safety regions for classification. First, for differentiating between safe and non-safe simulations, we achieve better and more efficient results compared to not using these topological features. The results were even more favorable for distinguishing between efficient and non-efficient simulations, ultimately yielding a very simple rule for differentiating between compliant and non-compliant simulations. These findings confirm the suitability of topological features for enhancing transparency and reliability in classification and safety detection tasks.
\section{Conclusions and future works}\label{sec:conclusion}

In this research, we presented an ML-based approach to compliant simulation, free of both collision and deadlock events, in mobile robot navigation. The proposed methodology contributes significantly to the field of eXplainable Artificial Intelligence (XAI) by integrating topological data analysis (TDA) and rule-based explanations to enhance interpretability in mobile robot navigation. The use of adjustable SVM classifiers and order statistics to define safety regions, followed by the extraction of Anchor rules, ensures that the decision-making process remains transparent and comprehensible. Overall, the proposed methodology achieves better and more efficient (simpler) results for collision avoidance compared to those obtained without using topological features as input in \cite{10.1007/978-3-031-63803-9_22}. Furthermore, we extend this approach to deadlock avoidance and, ultimately, we combine the safety regions and rules for both events, obtaining the corresponding safety region and rules for safe and efficient (compliant) simulations, avoiding these two negative events (collisions and deadlocks), obtaining accurate and promising results for that.
Beyond the immediate application in mobile robotics, the principles introduced in this work offer broader implications for xAI. The combination of topological insights with rule-based models presents a novel approach for generating explainable and robust decision boundaries in machine learning systems. In fields such as autonomous driving, industrial automation, and swarm robotics, where safety and efficacy are critical, the ability to construct interpretable safety regions could lead to more reliable and trustworthy AI-driven solutions.

While in this work the focus was to provide safety guarantees based on topological features, future research could explore more complex scenarios and behaviors, as well as incorporate additional topological parameters into the analysis. This could further enhance the quality of the generated safety regions, improving the interpretability, and optimizing the overall performance.

Finally, since the rules are obtained a posteriori to the simulations -—meaning that we need to have the complete simulation and calculate the topological features to determine whether the simulation falls within the safety margins defined by the obtained rules-—we leave as future work the definition of the simulation parameters a priori so that the resulting simulations exhibit a safe entropy value, that is, within the range identified as safe by the rules.

\subsection*{Code availability}
The code for the data and the experiments is available on a GitHub repository\footnote{\url{https://github.com/Cimagroup/Topological-Features-and-Explainable-Safety-Regions}}.

\subsection*{Acknowledgements} This work was partially supported by REXASI-PRO H-EU project (HORIZON-CL4-2021-HUMAN-01-01, Grant Agreement ID: 101070028), the Departmental Research Budget of the Department of Applied Mathematics I of Universidad de Sevilla, and the Future Artificial Intelligence Research (FAIR) project under the Italian National Recovery and Resilience Plan (Piano Nazionale di Ripresa e Resilienza - PNRR), Spoke 3 - ResilientAI.

%\addcontentsline{toc}{section}{References}

%Bibliography
\bibliographystyle{unsrt}  
\bibliography{biblio}

\end{document}